\documentclass[runningheads]{llncs}
\usepackage{graphicx}
\usepackage[utf8]{inputenc}
\usepackage{times}
\usepackage[T1]{fontenc}
\usepackage{graphicx}
\usepackage{todonotes}
\usepackage{multirow}
\usepackage{amsmath}
\usepackage{subfig}
\usepackage[export]{adjustbox}
\usepackage[square,numbers]{natbib}
\usepackage{tabularx,stackengine}
\newcolumntype{L}[1]{>{\minwd l{#1}}l<{\endminwd}}
\newcolumntype{C}[1]{>{\minwd c{#1}}c<{\endminwd}}
\newcolumntype{R}[1]{>{\minwd r{#1}}r<{\endminwd}}
\def\minwd#1#2#3\endminwd{\stackengine{0pt}{#3}{\rule{#2}{0pt}}{O}{#1}{F}{F}{L}}

\begin{document}
\title{Effect of depth order on iterative nested named entity recognition models}
%
%
%
\authorrunning{Wajsburt et al.}
%
\institute{
Sorbonne Université, Inserm, LIMICS, Paris, France
\and
SCAI, Sorbonne Université, Paris, France
}
\author{Perceval Wajsbürt\inst{1}, Yoann Taillé\inst{1,2}, Xavier Tannier\inst{1}\\
}

\maketitle              
\begin{abstract}
This paper studies the effect of the order of depth of mention on nested named entity recognition (NER) models. NER is an essential task in the extraction of biomedical information, and nested entities are common since medical concepts can assemble to form larger entities. Conventional NER systems only predict disjointed entities. Thus, iterative models for nested NER use multiple predictions to enumerate all entities, imposing a predefined order from largest to smallest or smallest to largest. We design an order-agnostic iterative model and a procedure to choose a custom order during training and prediction. To accommodate for this task, we propose a modification of the Transformer architecture to take into account the entities predicted in the previous steps. We provide a set of experiments to study the model's capabilities and the effects of the order on performance. Finally, we show that the smallest to largest order gives the best results.


\end{abstract}

\section{Introduction}

Biomedical concept recognition is a classical and essential task of natural language processing for biomedical applications~\cite{Ravanbakhsh2012AModels}, aiming to extract information such as symptoms, treatments, proteins, genes, dates, and durations from free text. Classic methods assume that entities are disjoint and formulate the problem as a sequence segmentation task, using word tagging schemes. However, in a real-world scenario, entities can compose or overlap, thus breaking the assumption that they are disjoint. For example, a temporal event "after anesthesia" contains the nested treatment entity "anesthesia."
\vspace{-5pt}
\begin{center}
    \includegraphics[width=33.8ex]{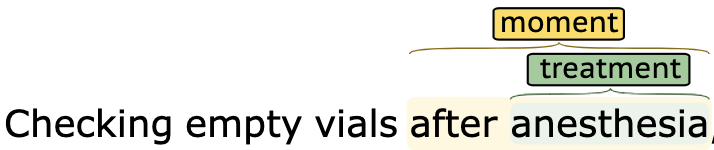}
\end{center}
\vspace{-5pt}
Multiple methods have been proposed to tackle this problem, ranging from exhaustive filtering of all possible spans to complex graph structures.

A class of methods deals with this problem of nested named entity recognition with a cascade of flat (non-nested) named entity recognition layers for different depths, i.e., predict the entities at a given depth iteratively starting either from large entities or short entities. The predictions of a given depth are used as additional input for the next prediction. We can argue whether the depth order matters during the training of such a model: is it easier for the model to predict large entities first and look inside its previous predictions for smaller ones, or predict small entities and compose them to build larger ones? To answer this question, we design an order-agnostic auto-regressive model based on the Transformer encoder architecture and a procedure to let it choose a custom order during training and prediction.

This work was originally designed to address Task 3 of the DEFT 2020 evaluation campaign (in French), and we further evaluated it on the classical GENIA dataset (in English). This DEFT task deals with the detection of named entities in texts describing clinical cases \cite{Grabar2018CAS:Cases}. The different types of entities are, on the one hand, pathologies and signs or symptoms (task 3.1), and on the other hand, \textit{anatomy}, \textit{anatomy examinations}, \textit{substances}, \textit{doses}, \textit{administration methods}, \textit{treatments (surgical or medical)}, \textit{values}, \textit{time}. More details about the challenge are presented in \cite{Cardon2020PresentationCases}, in which we describe our 3~official submissions. 

To our knowledge, our work is the first to evaluate a nested biomedical concept recognition system in the French language, which constitutes an additional contribution, as resources for languages other than English are very scarce~\cite{Neveol2016Clinical2016}.

\section{Related works}

Nested named entity recognition has been the subject of renewed attention in the biomedical NLP community since 2018, leading to many different approaches.

\paragraph{Exhaustive or semi-exhaustive methods} There are multiple approaches to tackle the task of nested named entity recognition.
One class of method addresses the problem by enumerating all possible spans of the input sequence and classify each one with its label, including a "no entity" class. \cite{Sohrab} classify each span independently by pooling its tokens reprentations. \cite{Xu} consider the left and right context when classifying the spans. \cite{Wangb} uses an LSTM cell \cite{Hochreiter1997LongMemory} to model dependencies between spans that differ by one token. \cite{Zheng} first filters candidate mentions by predicting all possible start and end tokens and then predicting a label for every mention that starts or end at one of the boundaries.

\paragraph{Layered methods} A different research direction focuses on layered models that iteratively predicts entity spans until none can be found. \cite{Fisher} uses a fixed number of layers and updates spans representations using a novel neural architecture. \cite{Ju} designs a layered architecture that predicts entities at each layer and merges the word representations before applying the next layer. \cite{Shibuya2019} computes tag scores for each word and decodes the spans by applying the Viterbi algorithm multiple times on a previously extracted subsequence, starting from the full sentence.

\paragraph{Hypergraph methods} Some methods model the span detection with hypergraphs to account for the non-linear structure of the tag sequences. \cite{Lu2015} design a CRF hyper-graph with various node types to model entity types and boundaries. However, cycles in the graphs of some samples required that the CRF normalization term had to be approximated, leading to a decreased performance \cite{Muis}. \cite{Muis} models the mention edges and transitions instead of solely modeling token tags. Their method, however, requires multiple graphs if more there is more than one entity type. Alternatively, \cite{Katiyar} only models mention tags and not their transitions, but allows a multi-label prediction for each token. They modify an LSTM layer to represent multiple states for a single token and perform decoding during the neural network execution.

\paragraph{Neural transition models} Transition models iteratively build the detected spans by chaining different actions. \cite{Wang} employs a set of three actions types (SHIFT, REDUCE-X, UNARY-X) that build a forest of binary word trees. \cite{Marinho2019} adds a new action (OUT) to allow building multiple mentions for a given span, especially fine-grained entity classes according to a predefined semantic hierarchy.

\paragraph{Other methods} \cite{Lin} uses Anchor Region Networks to first predict an anchor token that is contained by an entity and then predict the range of the mention from that anchor. This method leverage the insight that most mentions contain to a head token.

\paragraph{Language embedding} Pre-trained self-supervised language models \cite{Mikolov2013,Peters2018,Devlin2018} have improved downstream NLP tasks' performance. Most state-of-the-art methods use pretrained models as contextualized token features but do not alter the underlying model's structure. In contrast, we experiment with the augmentation of a pretrained transformer model to address the nested entity recognition task.

\section{Method}
\subsection{Model}

We detail here a new model to handle nested entity recognition. The model is an auto-regressive encoder-only Transformer \cite{Vaswani2017} taking as input a sequence of words and a list of entity mentions already extracted (empty list at the first iteration) and predicts as output a list of new mentions.
The entities predicted at each iteration do not overlap, but all the entities predicted at the end of the iterations may overlap.

\begin{figure}[t]
\centering
    \subfloat[\centering]{{\includegraphics[height=200pt, valign=c]{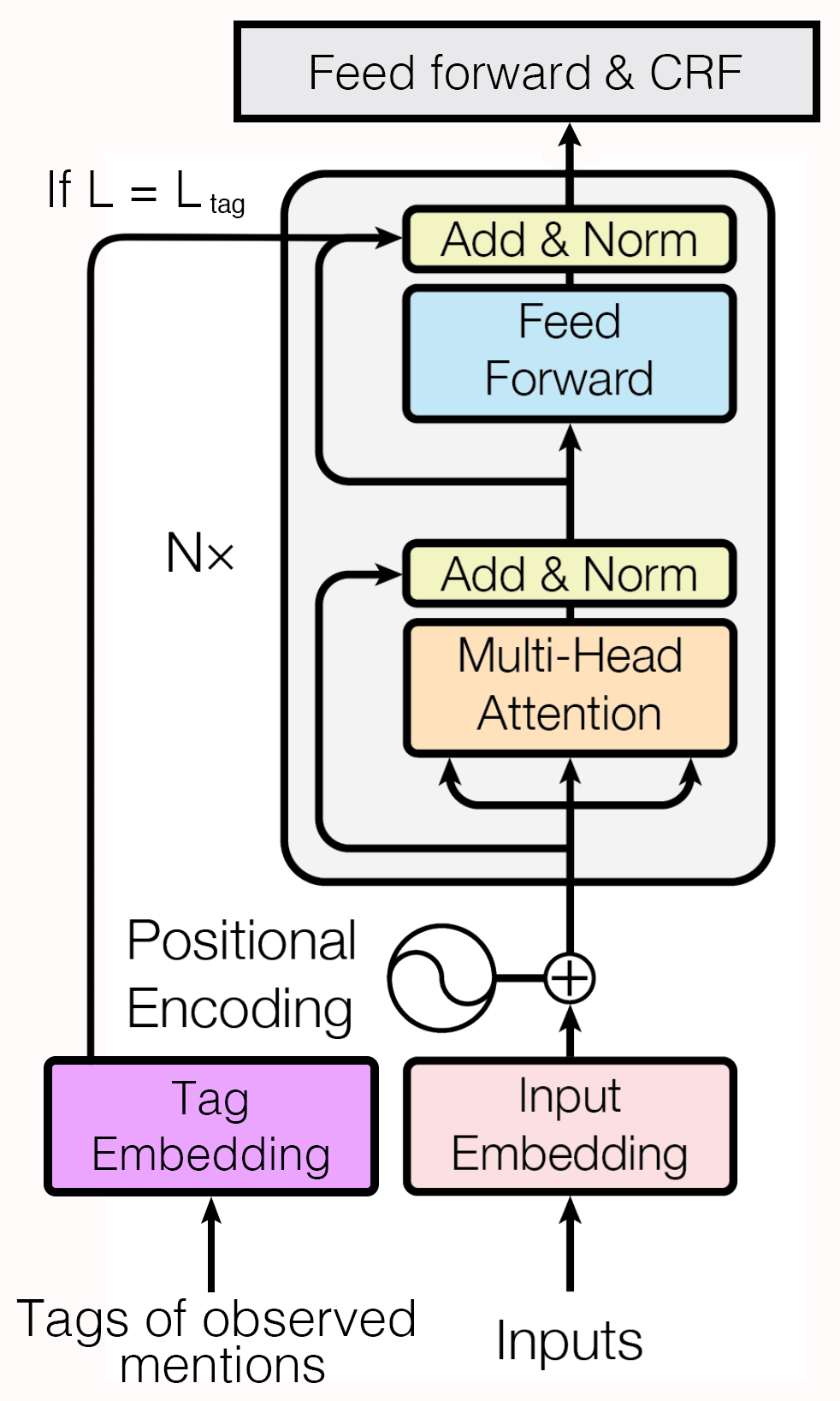}}}
\hspace{\stretch{1}}
    \subfloat[\centering]{{\includegraphics[height=200pt, valign=c]{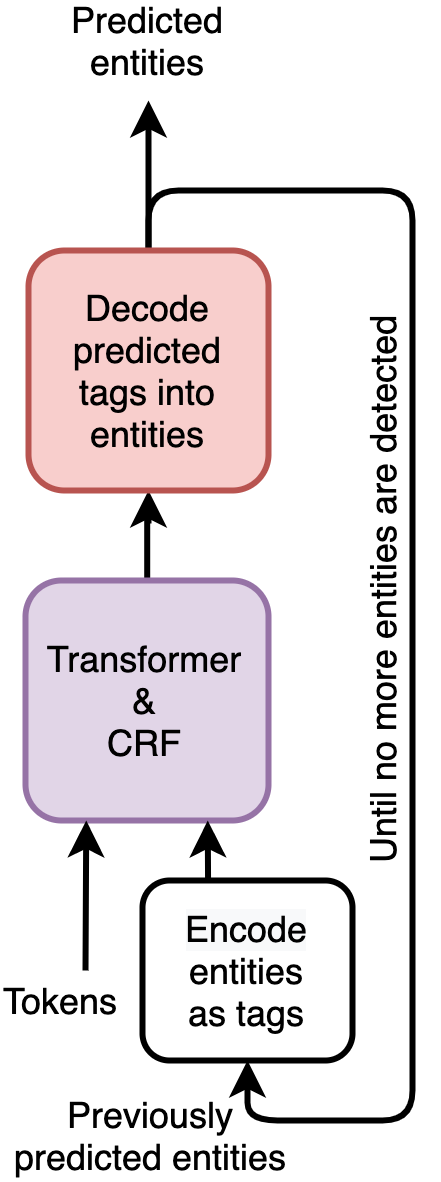}}\vspace{1in}}%
   \caption{(a) modified Transformer conditionnned on the tags of the previously observed mentions at the end of the layer $L_{tag}$ - figure adapted from \cite{Vaswani2017}; (b) Global model prediction diagram}
     \label{fig:model}
\end{figure}

We handle entity mentions in the form of tags assigned to each token with the conventional BIO or BIOUL formats \cite{Dai2015EnhancingTokenization,Ratinov2009DesignRecognition}, by embedding each tag into a multidimensional vector space. We then sum the tag embeddings of different entities (depths) at the same position with each other.
The sentences are tokenized and represented with the contextual embedding, Transformer model BERT (for English documents~\cite{Devlin2018}) or CamemBERT (for French~\cite{Martin2019}). The word embedding and the tag embeddings from previous iterations are fed into a linear CRF~\cite{Lafferty2001} predicting flat entity mentions (Figure~\ref{fig:model}a).

Each iteration of this model receives the tags predicted by the previous iteration as input to predict different mentions at each step (Figure~\ref{fig:model}b).

\subsection{Greedy order training}

During training, the goal is to maximize the probabilities assigned by the model to the lists of target entities. This goal comes with the difficulty of dealing with overlapping entities. 

We proceed in several steps and predict only non-interleaved entities at each run. However, several permutations, or valid prediction paths, lead to the same list of entities. For two nested statements
\vspace{-5pt}
\begin{center}
    \includegraphics[width=33.8ex]{./files/nested-example-1.png}
\end{center}
\vspace{-5pt}
(annotations that we call $T$ and $H$), we can predict $T$ first, then $H$ knowing $T$, or the opposite, i.e. choose to optimize between two objectives :

\begin{itemize}
    \item $P(T, H) =  P(T,H|T) \times P(T)$
    \item $P(T, H) =  P(T,H|H) \times P(H)$
\end{itemize}

In this symmetrical situation, a solution that would optimize both paths simultaneously, for example, by summing several tag losses for different depths, could lead to a non-optimal situation in which we ask the model to detect an incorrect combination of the two mentions. For example, these two non-overlapping but sub-optimal mentions:
\vspace{-5pt}
\begin{center}
    \includegraphics[width=35ex]{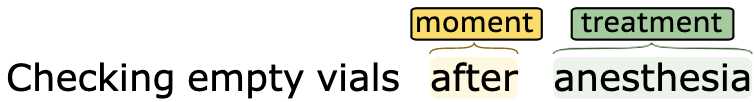}
\end{center}

Models like \cite{Ju} choose a strategy in advance (smaller mentions first, for example), but the risk is not to take advantage of all the interdependencies that make some mentions easier to find when you know the others.

Another solution is to choose the order of extraction leading to the model's best performance, measured in F-measurement. A greedy strategy is applied to select, among the non-overlapping combinations of mentions not observed in this batch, the closest to the mentions predicted in terms of F1 overlap. Intuitively, this means that a combination that is easier for the model to predict is preferred. During training, to simulate an extraction in progress, we randomly select in each sentence a subset of the entities and label them as already predicted entities.

\subsection{Inference}

For each sentence in the corpus, our model starts by predicting the most likely sequence of entities from the input token sequence alone since no mention has already been predicted. Then, we add them to the observed entities list and repeat the prediction until no more entities can be found.

\subsection{Model parameters}
We initialize the Transformer with CamemBERT \cite{Martin2019} weights for DEFT and BioBERT \cite{Lee2020BioBERT:Mining} for GENIA unless mentioned otherwise, and the remaining parameters using the method of \cite{Glorot2010UnderstandingNetworks}. Dropout \cite{Srivastava2014Dropout:Overfitting} is applied with a probability of 0.25 everywhere. We optimize the parameters by backpropagation with Adam \cite{Kingma2015} without weight decay, over 40 epochs for DEFT and 10 epochs for GENIA. We use two learning steps: one for the Transformer weights, initialized at $4\times10^{-5}$, and one for the rest of the model, initialized at $9\times10^{-3}$. The learning rate follows a linear decay schedule with a warmup for 10\% of the steps. We insert the tag embeddings in the Transformers at layer $L_{tag} = 6$ for BERT with 12 layers and 19 for BERT with 24 layers. On an Nvidia K80 GPU graphics card, learning on 100~documents takes about 20 minutes.

%
%

\section{Experiments and discussions}

\subsection{Datasets}

We conduct experiments on two nested named entity recognition datasets: DEFT~\cite{Cardon2020PresentationCases} and GENIA~\cite{Kim2003}, which present documents with different languages and different types and depths of entities (see Table~\ref{fig:dataset_stats}). For GENIA, we perform splits following \cite{Finkel2009}: the last 10\% of the sentences are used to test the model, the remaining 90\% are the training set. For DEFT, we used the provided train and test splits. In both cases, we split the training data into 90\% for training the model and 10\% for the development (validation) set. We selected the hyperparameters by grid search on the development set and trained on the training split for GENIA, and both training and development splits for DEFT.


\begin{table}[t]
    \centering
    \begin{tabular}{c|ccc|ccc}\hline
               & \multicolumn{3}{c|}{ GENIA (English)} & \multicolumn{3}{c}{ DEFT (French)} \\\hline
               & train & dev   & test  & train  & dev    & test  \\\hline
sentences      & 15022 & 1669  & 1855   & 1481   & 365   &  1024     \\
documents      & 1599  & 190   & 213    & 82     & 18    &  67     \\\hline
mentions       & 47027 & 4469  & 5600   & 6439   & 1498  & 4791      \\
mentions $D_0$ & 42965 & 4072  & 5007   & 5098   & 1226  & 3538      \\
mentions $D_1$ & 3959  & 394   & 1282   & 1282   & 261   & 1163      \\
mentions $D_2$ & 102   & 3     & 59     & 59     & 12    & 90      \\
mentions $D_3$ & 1     & 0     & 0      & 0      & 0     & 0      \\\hline
\end{tabular}
    \caption{Statistics of the datasets. $D_0$ are all the mentions that does not contain any mentions, $D_1$ are the mentions that contain one or more $D_0$ mention, etc.}
    \label{fig:dataset_stats}
\end{table}



\subsection{Baselines}


We compare our results against a simple flat NER model composed of a transformer and a CRF decoding layer that can only predict non-nested mentions. Since a choice is required during training as to which mentions should be predicted, we evaluate three modes: we only recover the shortest mention in a nested group, or only the largest, or let the model decide greedily.
We also compare our model against the state-of-the-art models on GENIA and the other participants' models on DEFT. 

\subsection{Results}
The results of our system and the baselines are presented in Tables~\ref{tab:genia-results} and \ref{tab:deft-results}. 

On the DEFT task 3.1, our model obtains the best F1 result of 0.66 (with the exact delimitation of mentions). On the DEFT task~3.2, the same model obtains a F1 of 0.778. Flat NER models lose between 10 and 20 points in F1, due to the large number of nested mentions. The 3.1 task containing longer entities, the flat large entities model, reaches the best performance of the flat models. Conversely, the flat short entities model obtains the best performance on the 3.2 task.
 
On the GENIA dataset, our best model reaches 0.7683 F1 with BioBERT large. We hypothesize that our method ranks lower on the latter dataset because it only uses BERT instead of BERT and other word features, and that the insertion of tags directly in BERT architecture may lead to loose some of the pretrained model abilities. We can also observe that flat NER is competitive with iterative models, which can be explained by the low ratio of nested mentions in the dataset.

\begin{table}[t]
    \centering
    \begin{tabular}{l|@{\hspace{0.05in}}c@{\hspace{0.05in}}|@{\hspace{0.05in}}c@{\hspace{0.05in}}|@{\hspace{0.05in}}c@{\hspace{0.05in}}}
\emph{GENIA dataset} & Precision & Recall & F1 \\\hline
\citet{Ju} & 0.785 & 0.713 & 0.747 \\
\citet{Wanga} & 0.780 & 0.702 & 0.739 \\
\citet{Wang} & 0.770 & 0.733 & 0.751 \\
\citet{Sohrab} & 0.932 & 0.642 & 0.771 \\
\citet{Lin} & 0.758 & 0.739 & 0.748 \\
\citet{Shibuya2019} & 0.763 & 0.747 & 0.755 \\
\citet{Strakova2019NeuralLinearization} (BERT, Flair) & & & 0.783 \\
\citet{Wangb} (BERT, Flair) & 0.803 & 0.783 & 0.793 \\\hline
Flat short entities (BERT base) & 0.7929 & 0.6975 & 0.7422 \\
Flat large entities (BERT base) & 0.8149 & 0.7070 & 0.7571 \\
Flat greedy (BERT base) & 0.8141 & 0.7095 & 0.7582 \\
(our) greedy (BERT base) & 0.8126 & 0.7211 & 0.7641 \\
(our) large-to-short (BERT base) & 0.8016 & 0.7184 & 0.7577 \\
(our) short-to-large (BERT base) & 0.8028 & 0.7336 & 0.7666 \\
(our) short-to-large (BERT large) & 0.7933 & 0.7448 & 0.7683 \\
\end{tabular}
    \caption{GENIA test performance (systems reaching performance above 0.74 F1)}
    \label{tab:genia-results}
\end{table}

\begin{table}[t]
    \centering
    \begin{tabular}{l|c|c|c||c|c|c||c}
        & \multicolumn{3}{c||}{ DEFT 3.1 } & \multicolumn{3}{c||}{ DEFT 3.2 } & Global \\\hline
        & P & R & F1 & P & R & F1 & F1 \\\hline
        HESGE (BERT large) & 0.702 & 0.624 & 0.660 & 0.788 & 0.725 & 0.755 & \\
        Median DEFT & & & 0.4557 & & & 0.6151 & \\\hline
        Flat short (BERT base) & 0.6093 & 0.2277 & 0.3315 & 0.7570 & 0.6690 & 0.7103 & 0.6228 \\
        Flat large (BERT base) & 0.6093 & 0.6076 & 0.6085 & 0.7223 & 0.3152 & 0.4389 & 0.5094  \\
        Flat greedy (BERT base) & 0.6082 & 0.3481 & 0.4428 & 0.7848 & 0.6046 & 0.6830 & 0.6201  \\
        (our) greedy & 0.6263 & 0.6090 & 0.6175 & 0.7615 & 0.7416 & 0.7514 & 0.7134 \\
        (our) large to short & 0.6257 & 0.6062 & 0.6158 & 0.7407 & 0.7470 & 0.7439 & 0.7082 \\
        (our) short to large & 0.6105 & 0.6194 & 0.6149 & 0.7564 & 0.7449 & 0.7506 & 0.7120 \\
        (our) short to large (BERT large) & 0.6613 & 0.6595 & 0.6604 & 0.7806 & 0.7759 & 0.7783 & 0.7447 \\
    \end{tabular}
    \caption{DEFT test performance}
    \label{tab:deft-results}
\end{table}

\subsection{Ablations}

\subsubsection{Tag embeddings}

\begin{figure}[t]%
    \centering
    \subfloat[\centering GENIA]{{\includegraphics[height=115pt]{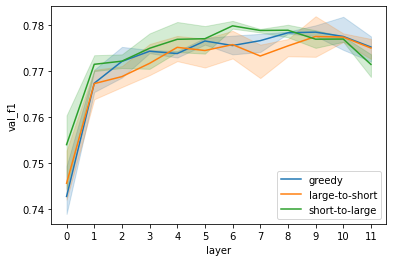}}}
    \subfloat[\centering DEFT]{{\includegraphics[height=115pt]{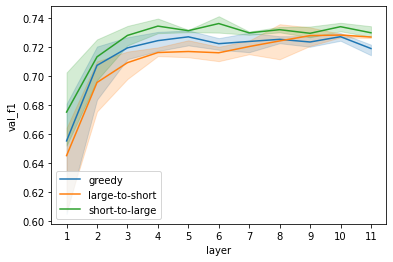}}}%
    \caption{Model performance on the GENIA and DEFT validation datasets wrt different insertion layer indices of the observed tag embeddings and training depth orders. Each point is the average over 3 runs. We only plotted the results from the second layer for DEFT to improve readability, all models reaching around 0.50 F1 when layer = 0.}
    \label{fig:tag-layer-index-genia}%
\end{figure}

\paragraph{Insertion layer} We study the effect of different tag embeddings configurations on the system's performance. We can see in Figure \ref{fig:tag-layer-index-genia} the validation f1 metrics of the system against the index of the BERT layer at which we add the embeddings. The performance decreases in the last layers since the system has no "time" to take the previous predictions into account. However, we also observe a maximum at layer 6 and lower performance in earlier layers.

\paragraph{Tag scheme} We analyze the performance of two common tag schemes: BIO (Begin, Inside, Outside) and BIOUL (BIO with Unary and Last tags) to encode observed (i.e., previously predicted entities). Results can be found in Table \ref{fig:bio_bioul}. The BIOUL tag scheme shows better overall results than the BIO scheme. This conclusion is similar to what others \cite{Lample2016} have observed for flat named entity recognition. However, the BIOUL "reading" tag scheme's better results show that the system can better use previous entities for its predictions when given begin and end bounds.

\begin{table}
    \centering
    \begin{tabular}{l|c|c}\hline
               & write BIO              & write BIOUL  \\\hline
read BIO       & \,0.7221 ($\pm 2.98e^{-3}$)\,	& \,0.7341 ($\pm 6.09e^{-3}$)\, \\
read BIOUL     & \,0.7261 ($\pm 5.40e^{-3}$)\,	& \,\textbf{0.7368} ($\pm 4.28e^{-3}$)\, \\\hline
\end{tabular}
    \caption{Performance of the BIO and BIOUL reading and writing tag schemes on the DEFT validation dataset.}
    \label{fig:bio_bioul}
\end{table}

\subsubsection{Depth prediction order}
We study the effect that forced prediction order during training has on model performance. We compared three prediction modes: top to bottom, bottom to top, and greedy decoding.
In the top to bottom mode, given a previously predicted entity at depth D, we force the model to predict a named entity located at depth D+1.
In the bottom to top mode, we use the inverse depth as training order. Finally, in greedy decoding mode, we let the model choose the mentions by selecting those closest to its prediction. 

From Figure \ref{fig:tag-layer-index-genia} we can observe that the short-to-large training order obtains the highest performance on both GENIA and DEFT validation splits. The large-to-short depth training order obtains the lowest accuracy. We hypothesize that learning to detect the smallest, and often easier, entities first leads the model to learn how to compose new entities from small entities. On the other hand, learning to predict large, and often more difficult, mentions first, must lead the model to overfit on these large mentions and fail to recover smaller nested mentions when the largest ones are wrongly predicted. The greedy training reaches an intermediate performance, so we conclude that a learned prediction order is suboptimal.

\section{Conclusion}

This paper proposes an architecture to perform named entity recognition based on iterative predictions and dynamic mention matching during training. We also provided insights into the model behavior and showed that training depth mention order impacts performance on auto-regressive layered named entity recognition models, and short-to-large order obtains the best results. 

\bibliographystyle{splncs04nat}
\bibliography{references}
%
%
%
%
%
\end{document}